**Fully Automated Myocardial Infarction Classification using Ordinary Differential Equations**

Getie Zewdie[1],Momiao Xiong[1]

[1]The University of Texas School of Public Health, Division of Biostatistics, Houston, Texas 77030, USA

**Running title:** Fully Automated Myocardial Infarction Classification

[*]Address for correspondence and reprints: Dr. Momiao Xiong, Human Genetics Center, The University of Texas Health Science Center at Houston, P.O. Box 20186, Houston, Texas 77225, (Phone): 713-500-9894, (Fax): 713-500-0900, E-mail: Momiao.Xiong@uth.tmc.edu.

**Abstract**.

Portable, Wearable and Wireless electrocardiogram (ECG) Systems have the potential to be used as point-of-care for cardiovascular disease diagnostic systems. Such wearable and wireless ECG systems require automatic detection of cardiovascular disease. Even in the primary care, automation of ECG diagnostic systems will improve efficiency of ECG diagnosis and reduce the minimal training requirement of local healthcare workers.  However, few fully automatic myocardial infarction (MI) disease detection algorithms have well been developed. This paper presents a novel automatic MI classification algorithm using second order ordinary differential equation (ODE) with time varying coefficients, which simultaneously captures morphological and dynamic feature of highly correlated ECG signals. By effectively estimating the unobserved state variables and the parameters of the second order ODE, the accuracy of the classification was significantly improved. The estimated time varying coefficients of the second order ODE were used as an input to the support vector machine (SVM) for the MI classification. The proposed method was applied to the PTB diagnostic ECG database within Physionet. The overall sensitivity, specificity, and classification accuracy of 12 lead ECGs for MI binary classifications were 98.7%, 96.4% and 98.3%, respectively. We also found that even using one lead ECG signals, we can reach accuracy as high as 97%. Multiclass MI classification is a challenging task but the developed ODE approach for 12 lead ECGs coupled with multiclass SVM reached 96.4% accuracy for classifying 5 subgroups of MI and healthy controls.

**Introduction**

Myocardial infarction commonly known as heart attack is one of the heart diseases and remained the major causes of death and disability worldwide [1]–[4]. According to the national vital statistics [2], about 715,000 American have heart attack every year. Myocardial infarction occurs when myocardial ischemia or chest pain, a reduced blood supply to the heart, exceeds a critical threshold. Myocardial ischemia is an imbalance between blood supply and demand to the heart through the two major coronary arteries. When sufficient amount of blood does not flow to the heart properly, the heart muscles get injured or damaged due to lack of enough oxygen. This cell injury or damage because of lack of enough oxygen is commonly known as myocardial ischemia and prolonged myocardial ischemia will develop into myocardial infarction unless proper medication is used [5]

An ECG signal is a graphic electrical signal generated by the heartbeats. The cell's membrane potential called depolarization is associated with electrical changes resulted from the contraction of body muscles. When an individual is fully relaxed and no skeletal muscles are contracting, the electrical changes associated with the contraction of the heart muscle will be apparently perceived and electrodes attached to the surface of the body can detect the depolarization. Thus, the appropriate processing of ECG signals and effectively detecting the signals information is very crucial as it determines the condition of the heart [6]–[9]. The ECG signals have been widely used for detection and diagnosis of MI [6], [7].  However, the ECG signals are examined manually by cardiologists or specialized device only available in large hospitals [11]. Visual inspection of the whole ECG trend even for one individual is cumbersome. Moreover, cardiologists may miss some important information when they visually examine ECG signals and which eventually may lead to wrong interpretation of ECG signals.  This will cause problems particularly in low and middle-income developing countries where well-trained healthcare workers and clinics with rich resources are less available. Therefore, in clinical applications there is urgent need to develop automated MI classification algorithms for ECG data analysis to improve the diagnosis accuracy and reduce the training of cardiologists. In the last few years, we have observed the development of wearable and wireless ECG system and a shift of healthcare from hospital-centered to more individual centered mHealth system [8]. Manual examination of ECG signals for real time mobile wireless ECG monitoring systems are inappropriate and infeasible.  Real time ECG monitoring also demands to develop novel algorithms for automatic MI classification using ECG signals. There had been a growing interest in the automatic ECG classification and several classification approaches (discriminant analysis, support vector machine, neural network, decision trees, etc.) were developed [9]–[18]. Although these developed approaches can automatically analyze ECG signals without manual interference from cardiologists, they did not provide the desired optimal classification quality due to the lack of ability to extract useful information from ECG morphology and ECG dynamic features. Recently, some authors take both inter-ECG differences and intra-ECG differences (for example, normal and abnormal heartbeats) as features and have improved MI classification [19].  However, heartbeats are labeled manually. Therefore, ECG analysis for MI classification is not fully automated. To automatically analyze ECG data for MI classification while still using both inter and intra-ECG differences as selected features, Sun et al. (2012) [19] developed a latent topic multiple instance leaning (LTMIL) method where normal and abnormal heartbeats were identified by cluster analysis. LTMIL has improved MI classification accuracy. But the performance of LTMIL depends on two parameters: number of clusters and the scale factor μ. Automatic parameter tuning in the LTMIL is still a challenging task. This paper serves three purposes. The first purpose is to develop fully automated ECG analysis system for MI binary classification, which can also substantially improve the accuracy of MI diagnosis. MI can be divided into multiple subgroups. To our knowledge, very few papers in the literatures have investigated MI multiclass classification. Therefore, the second purpose of the paper is to develop fully automated ECG system for MI multiclass classification and to evaluate their diagnosis accuracy. To facilitate their widespread use, wireless ECG sensor for real time monitoring must be very comfortable to wear. Conventional 12-lead ECG system is too complex to be employed in ECG real time monitoring system. Designing a wearable and wireless ECG system for home application requires as few leads as possible. The third purpose is to evaluate the performance of the proposed method using single lead or few leads for MI classification and study the feasibility of a single lead wearable and wireless ECG monitor for MI diagnosis. To achieve these goals, second order ordinary differential equations (ODEs) were proposed to automatically detect myocardial infarction from ECG signals. Differential equations have been used broadly in Biology, Chemistry, Physics, Economics, Engineering, and in Medicine [20]–[26]. It has been also used in other areas like in Psychology for self-regulating systems [27]. Differential equations can describe almost any systems that undergoing change [21], [28]–[30]. Mathematician and statisticians have studied the natures of differential equations for several decades and developed different techniques for solving differential equations [24], [31]–[34]. However, most investigations have mainly focused on the solutions of ODE, only few papers have been devoted to estimate the parameters of ODE. In this paper, a two-step ODE-based MI classification method will be illustrated. First, we developed a method to estimate the time varying coefficients of second order ODE that models the ECG signals. Second, once the time varying coefficients of the second order ODE for modeling ECG are estimated, the maximum values of the time varying coefficients in the interval for each ECG were taken as inputs to the support vector machine for the ECG classification. The results show that the dynamic approach is an efficient method for dimension reduction. The developed auto MI binary and multiclass classification system has a significant improvement in terms of sensitivity and specificity for the detection of MI compared to the existing methods. The results also demonstrate that using a single lead ECG records may reach high classification rates and that developing a single lead wearable and wireless ECG monitor is feasible.

## Second order ODE for ECG analysis

### A. The second order ODE with time varying coefficients for modeling ECG

Second order ODE was used to model the dynamic system of the ECG signals, which has non-stationary trends. ECG signals are viewed as state variables that determine the dynamics of autonomous heart system. We usually use homogeneous ODE without forces in the right side to model dynamics of the ECG. Let $x(t)$ be a state variable that determines an ECG signal at time $t$. Since the second order time varying ODE is a simple, but also very general differential equation for modeling dynamics of the biological systems, we assume that the state variable $x(t)$ of ECG satisfies the following second order homogeneous ODE with time

varying coefficients:

$$\frac{d^2 x(t)}{dt^2} + b_1(t)\frac{dx(t)}{dt} + b_0(t)x(t) = 0 , \qquad (1)$$

Where $b_1(t)$ and $b_o(t)$ are time varying coefficients of the ODE. The curve of the ECG signals of each lead or channel was modeled by the second order ODE. Therefore, if we observe 12 leads ECG signals for each individual, then 12 second ODEs with different time varying coefficients will be used to model the state variables that determine all 12 leads ECG recordings. The time varying coefficients of the differential equations were a summarization of the ECG signals of each channel. Thus, all the grid points of the ECG signals for every channel and over 100,000 features were summarized by the two coefficients of the differential equations.

## B. Estimation of the Unobserved State Variables

The state variables of the heart dynamic systems are unobservable. We can only observe their measurements, ECG recordings. ECG recordings include the measurement errors and noises of the heart dynamic system. We need to estimate state variables from the observed ECG recordings with measurement errors and noises. Consider $n$ individuals sampled at $T$ time points. State variable of the i[th] individual at j[th] time point can be modeled by:

$$y_i(t_j) = x_i(t_j) + \varepsilon_i(t_j), \quad i = 1, \dots, n, j = 1, \dots, T, \qquad (2)$$

Where $y_i(t_j)$ is the observed ECG recording at time $t_j$ and let $\varepsilon_i(t_j)$ be independent with mean zero and common variance $\sigma^2$.

Applying Taylor expansion:

$$\begin{aligned}
x_i(t_j) &= x_i(t_0) + (t_j - t_0)x_i^{(1)}(t_0) + \dots + \frac{(t_j - t_0)^p x_i^{(p)}(t_0)}{p!} \\
&= Z(t_0)_{p,j}^T \alpha_i(t_0),
\end{aligned} \qquad (3)$$

Where,

$$Z(t_0)_{p,j} = \left[1, (t_j - t_0), (t_j - t_0)^2, (t_j - t_0)^3, \dots, (t_j - t_0)^p\right]^T, \text{ and}$$

$\alpha_i(t_0) = [x_i(t_0), x_i^{(1)}(t_0), ..., x_i^{(p)}(t_0)]^T$.

Let

$Y_i = [y_i(t_1), ..., y_i(t_T)]^T$ , $Z_p(t_0) = [Z(t_0)_{p,1}, ..., Z(t_0)_{p,T}]^T$.

For convenience, we can write $Z_p(t_0)$ as:

$$Z_p(t_0) = \begin{bmatrix} 1 & (t_1 - t_0) & (t_1 - t_0)^2 & ... & (t_1 - t_0)^p \\ 1 & (t_2 - t_0) & (t_2 - t_0)^2 & ... & (t_2 - t_0)^p \\ 1 & (t_3 - t_0) & (t_3 - t_0)^2 & ... & (t_3 - t_0)^p \\ \vdots & \vdots & \vdots & ... & \vdots \\ 1 & (t_T - t_0) & (t_T - t_0)^2 & ... & (t_T - t_0)^p \end{bmatrix}$$

We can apply the weighted least square techniques to estimate the unobserved state variables. We minimize the following equation with respect to $\alpha_i(t_0)$.

$$\min_{\alpha_i(t_0)} \quad (Y_i - Z_p(t_0)\alpha_i(t_0))^T W_{hi}(Y_i - Z_p(t_0)\alpha_i(t_0)), \qquad (4)$$

Where,

$$W_{hi} = \begin{bmatrix} w_{hi}(t_1) & 0 & 0 & ... & 0 \\ 0 & w_{hi}(t_2) & 0 & ... & 0 \\ 0 & 0 & w_{hi}(t_3) & ... & 0 \\ \vdots & \vdots & \vdots & ... & \vdots \\ 0 & 0 & 0 & ... & w_{hi}(t_T) \end{bmatrix}$$

$w_{hi}(t_j) = \dfrac{K(\dfrac{t_j - t_0}{h_i})}{h_i}$ and $K(.)$ is a kernel function.

Solving the optimization problem (4) by weighted least squares, we obtained

$\hat{\alpha}_i(t_0) = (Z_p(t_0)^T W_{h_i} Z_p(t_0))^{-1} Z_p(t_0)^T W_{h_i} Y_i$.

## C. Parameter Estimation of Second Order ODE

After we estimate the state variables, we are in a position to estimate the time varying coefficients in the second ODE. First, we substitute a smoothing estimate of $x(t)$ in model (1) discussed in Section B and then use the least squares method to estimate the time varying coefficients of the ODE. Consider second order ODE of the form:

$$\frac{d^2x_i(t)}{dt^2} + \theta_2(t)\frac{dx_i(t)}{dt} + \theta_1(t)x_i(t) = 0, i = 1, 2, \ldots, n \qquad (5)$$

We can again use Taylor expansion to approximate the parameters $\theta_k(t), k = 1, 2$

$$\theta_k(t_j) = \theta_k(t_0) + \theta_k'(t_0)(t_j - t_0)$$

$$= z_j^T \beta_k(t_0) \qquad (6)$$

Where

$z_j = [1, t_j - t_0]^T$ and $\beta_k(t_0) = [\theta_k(t_0), \theta_k'(t_0)]^T$.

Substituting the estimated state variable and its derivatives, and equation (6) into equation (5), we obtain

$$x_i^{(2)}(t_j) + z_j^T \beta_2(t_0) x_i^{(1)}(t_j) + z_j^T \beta_1(t_0) x_i(t_j) = 0 \qquad (7)$$

Define,

$R_i(t_j) = [x_i^{(1)}(t_j), x_i(t_j)]$ and $\beta(t_0) = [\beta_2^T(t_0), \beta_1^T(t_0)]^T$.

Now, we can write equation (7) as:

$$x_i^{(2)}(t_j) + [x_i^{(1)}(t_j)z_j^T, x_i(t_j)z_j^T]\beta(t_0) = 0, i = 1, \ldots, n. \qquad (8)$$

Or we can write it as:

$$x_i^{(2)}(t_j) + [R_i(t_j) \otimes z_j^T]\beta(t_0) = 0 \qquad (9)$$

Where $\otimes$-denotes Kronecker product. To estimate the parameters $\hat{\beta}(t_0)$, we minimize the following equation via least squares technique:

$$\sum_{i=1}^{n} \sum_{j=1}^{T} \left[ x_i^{(2)}(t_j) + [R_i(t_j) \otimes z_j^T]\beta(t_0) \right]^2 \qquad (10)$$

The estimated parameters are then given by

$$\hat{\beta}_0(t) = -\left\{ \sum_{i=1}^{n} \sum_{j=1}^{T} [R_i^T(t_j)R_i(t_j) \otimes z_j z_j^T] \right\}^{-1} \sum_{i=1}^{n} \sum_{j=1}^{T} R_i^T(t_j) \otimes z_j) x_i^{(2)}(t_j) \quad (11)$$

## Results

## A. Data description

To evaluate the performance of the proposed dynamic method for ECG_based MI classification and compare our results with others, we used the PTB diagnostic ECG database within Physionet, which has the largest number of ECG recordings related to myocardial infarction and were used by Sun et al [19]. There were 148 subjects with myocardial infarction, 18-cardiomyopathy/heart failure, 15 bundle, branch block, 14 dysrhythmia, 7-myocardial hypertrophy, 6-valvular heart disease, 4 myocarditis, 4 miscellaneous, and 52 healthy controls (HC). Clinical summary was not available for 22 subjects. There were 209 men and 81 females with ages ranging from 17 to 87. A total of 522 ECG records were included in the analysis. Specifically, we used 368 ECG records with MI, 80 ECG records with HC, 20 ECG records with cardiomyopathy/heart failure (CH), 17 ECG records with bundle branch block (BBB), 16 ECG records with dysrhythmia (DY), and 21 ECG records with myocardial hypertrophy, valvular heart disease, myocarditis, and miscellaneous combined (VHD) in this analysis. One to fiverecords represented each subject and each record includes 15 simultaneously measured ECG signals. The 15 ECG signals were the conventional 12 leads (i, ii, iii, avr, avl, avf, v1, v2, v3, v4, v5, v6) plus the three Frank lead ECGs (vx, vy, vz). Each signal was digitalized at 1,000 samples per second, with 16-bit resolution over a range of $\pm$ 16.384mV. The PTB diagnostic ECG database was used for MI classification [35], [36].

## B. Overall workflow of classification analysis

The block diagram of the dynamic model for MI classification from ECG signal is shown in Figure 1. The data analysis is divided into three steps: (1) ECG data preprocessing, (2) the estimation of parameters in the second order ODE, and (3) classification. Wave Form Data Base (WFDB) software was used access to the physionet database and downloads all PTB diagnostic datasets. Statistical analysis was performed using the statistical packages R version 3.0.2 (R Foundation for Statistical Computing, Vienna, Austria). Two R packages (dct and fttw) were used for discrete cosine transforms and an ECG signal preprocessing. We also used the R package desolve for solving initial value problem of ordinary differential equations. On average, we had over 100,000 features or data points for each lead of ECG records. Each ODE has two coefficient functions as parameter functions to be estimated. Therefore, we had the same number of time varying coefficient parameter estimates from the second order ODE. For the two-coefficient function estimates of the ODE, we took the maximum value from all the time varying coefficient estimates of each coefficient function as features for each lead of ECG curve. As a consequence, each lead of ECG had two features. We used them as inputs for the classification. Support vector machine (SVM) was used as classifier [5]. To unbiasedly evaluate the performance of the proposed dynamic model for MI classification, a 10 fold cross validation (CV) was used to calculate the average sensitivity, specificity and classification accuracy. Specifically, the original sample is randomly partitioned into 10 equal size subsamples. Of the 10 subsamples, a single subsample is retained as the test dataset, and the remaining 9 subsamples are used as training datasets. The cross-validation process is then repeated 10 times (the folds), with each of the 10 subsamples used exactly once as the test dataset. The 10 results from the folds can then be averaged to produce a single estimation of the sensitivity, specificity, and classification accuracy.

## C.  Performance evaluation

For binary classification, sensitivity, specificity and classification accuracy were used to measure performance of binary classification tests. Sensitivity of the classifier is defined as the proportion of correctly classified patients; specificity is defined as the proportion of the correctly classified healthy individuals, and classification accuracy is defined as the proportion of correctly classified individuals (patients are correctly classified as patients and healthy individuals are correctly classified as healthy). Since multiclass classification involves more than two categories, it is difficult to define specificity. The traditional definition of specificity works only for binary classification. We only use sensitivity that is defined for each class and total classification accuracy to measure the performance of multiclass classifier. The sensitivity of ith class is defined as the proportion of the correctly classified individuals in the ith class. Total classification accuracy is defined as the proportion of correctly classified individuals.

## D. Binary Classification Between MI ECGs and HC ECGs

We first study classification between ECGs with MI and ECGs with HC. A total of 368 ECG records with MI

and 80 HC ECG records were used for MI classification. Each ECG curve contained the sampled ECG signals at 115, 000 time points.  For each lead of each individual, we used the second order ODE with time varying coefficients to fit the ECG curves. The maximum of the estimated differential equation coefficient functions  over the sampled interval of ECG curve from the second order time varying ODE were used as input variables for the MI classification. For each individual with 12 leads of ECG records, we had twenty-four estimated features, each lead of ECG records had two feature.

The illustration for the accuracy of the approximation of the second order time varying ODE to the observed ECG signals is depicted on Figure 2. We observed from Figure 2 that the second ODE approach outperformed the most commonly used approach to curve fitting, the cubic spline approach. Figure 3 showed that the maximum value of the time varying coefficients of the second order ODE for the ECG curves of lead V6 can be used to well discriminate between the ECG records with MI and the ECG records with HC. The proposed dynamic method was first used to discriminate between 368 ECGs with MI and 80 HC ECGs. We used tenfold cross validation to evaluate the performance of the dynamic method for MI classification. The average sensitivity, specificity and accuracy of the MI classification over the ten folds in the training and test datasets using each of 12 leads, three leads (I, II, III) and all 12 leads were summarized in Table 1. Several remarkable features emerged from Table 1. First, even using ECG curve of one lead, we can obtain high classification accuracy. The highest and lowest classification accuracy in the test dataset by ECG curve of one lead was 97.01% (lead III) and 95.31% (lead Avr), respectively. Second, using ECGs from three leads I, II and III, has improved classification accuracy. The sensitivity, specificity, and total accuracy of the three combined leads were 98.2%, 95.8%, and 97.8%, respectively. Third, if we combined the 12 standard leads together, we reached sensitivity, specificity, and classification accuracy as high as 98.7%, 96.4%, and 98.9%, respectively. It is interesting to note that application of the recently developed latent topic multiple instance learning (LTMIL) method [19] to the same dataset could only reached sensitivity 92.3% and specificity 88.1% without using tenfold cross validation, which was much lower than the results of the dynamic method.

**E. Multiclass Classification**

MI can be divided into multiple subgroups in clinical diagnosis, depending on where infarction takes place in ECGs. To assess the power of the dynamic method for MI multiclass classification, we partitioned ECGs of MI in the PTB diagnostic ECG database into five groups: MI, CH, BBB,DY, and VHD.

There are two types of approaches to multiclass SVM. One is to transform a multiclass classification problem into a binary classification problem (One-against-Rest, One-against-One) by combing several binary classifiers while the other is to simultaneously predict multiple classes by directly considering all data in one optimization formulation. Here, direct multiclass SVM was used to classify multiple groups of MI ECGs. Figure 4 plotted the patterns of the maximum of the time varying coefficients $\beta_0(t)$ and $\beta_1(t)$ of the second ODE over the sampled interval of the ECG which can well discriminate six subgroups.

The results of multiclass SVM for classifying five groups of MI and HC were summarized in Tables 2 and 3. In general, using ECGs of 12 leads, the dynamic method can predict five subgroups of MI and healthy controls (HC) with high accuracy (Total average accuracy for correctly classifying five subgroups of MI and HC in the test datasets was 96.37% and average accuracy for prediction of VHD, DY, BBB or CH was 90.1%, 91.8%, 93.5% and 96.3%, respectively). The proposed approach substantially outperformed LTMIL [19]. Either single leads or 12 leads had high successful rates to predict MI and HC. But, the ability of single leads to predict VHD, DY, BBB and CH varied. The performance of AVF, AVL, AVR, I and II for prediction of DY, BBB and CH was poor. However, to our surprise, precision of prediction of VHD, DY, BBB and CH by each of V1, V2, V3, V4, V5 and V6 leads was high. For example, average accuracy of prediction of VHD, DY, BBB and CH by lead V6 was 78.2%,81.2%, 84.2% and 90.9%, respectively.

**Conclusion**

The purpose of this paper is to develop an efficient method for ECG signal analysis and automatic detection of MI from patients' ECGs and address several essential issues in devising fully automated MI classification system from ECGs for aiding healthcare workers in making a correct diagnostic decision of MI. The first issue for the ECG signal analysis and automatic classification of MI is how to capture dynamic and morphological features of MI from ECG signals with more than 100,000 time points. Selection of a number of informative features from many irrelevant and redundant variables is a challenging task, but also is a key to the success of MI classification. We developed a second order time-varying ODE as a powerful tool to extract features of MI from ECGs. We showed that all inherent dynamic and morphological information of MI in ECGs were implied in the time-varying parameters of the second ODE and that using the maximum parameters of $\beta_0(t)$ and $\beta_1(t)$ of the ODE in the whole sampled time interval as the selected features dramatically reduced features from more than 100,000 to two. These features were automatically extracted and no any additional manual procedures were needed. The features selected by the second ODE coupled with the standard SVM substantially outperformed the existing methods.

The second issue is the performance of ECGs from one lead for MI classification. In the past several years, the wearable wireless ECG monitoring system has rapidly been developed. Using mobile ECG monitoring systems is of paramount significance for early detection of cardiovascular disease [8]. However, to enable their widespread use, wireless ECG sensor must be very comfortable to wear. Conventional 12-lead ECG system is too complex to be employed in ECG real time monitoring system. Designing a wearable and wireless ECG system for home application requires as few leads as possible. However, an essential question is how many leads should be used to ensure the accuracy of MI classification. In this paper, we showed that even if one lead was used, the sensitivity, specificity and accuracy were reached as high as 97.5%, 94.7% and 97%, respectively. This result implies that it is highly likely to develop wireless single-pad ECG systems with satisfied classification accuracy.

The third issue is multiclass classification. MI is divided into multiple subgroups. Identify subgroups of MI will provide useful information for monitoring progression of MI and personal medicine. However, multiclass classification is a challenging task. In general, a multiclass classifier has lower classification accuracy than a binary classifier. Can we develop an accurate and automatic MI multiclass classification system? In this paper, we showed that the proposed multiclass MI classification system reached as high as 96.37% accuracy for classifying five subgroups of MI and HC group. This demonstrated that the second order ODE coupled with multiclass SVM could provide accurate and automatic MI multiclass classification. The results in this report are preliminary.  Our intension is to stimulate further discussions regarding the study designs and analytical methods for developing automatic offline and real time ECG classification system to improve early detection of MI and reduce the mortality rate of heart disease.

## ACKNOWLEGGMENTS

Financial support for this study was provided by Grant 1R01AR057120–01,1R01HL106034-01,and 1U01HG005728-01 from the National Institutes of Health.

**Figure Captions:**

**Figure 1**. Overall workflow of MI classification from ECG signals

**Figure 2.** Comparing the performance of ordinary second order differential equation vs. cubic spline smoothing

**Figure 3**. Support vector machine binary classification of MI from ECG signals using ordinary second order differential equations

**Figure 4**. Support vector machine Multiclass classification using ordinary second order differential equations

**Figure 5**. Box plot for all the 12 channels (leads): For individual one with myocardial infarction and another individual from the healthy control. (The two individuals from the two groups are chosen arbitrarily to see the mean, median, and the percentiles)

**Table 1. Tenfold cross validation results for MI versus the healthy control**

|  | Training Dataset | | | Test Dataset | | |
|---|---|---|---|---|---|---|
|  | Sensitivity | Specificity | Accuracy | Sensitivity | Specificity | Accuracy |
| Avf | **0.989** | **0.966** | **0.9849** | **0.972** | **0.926** | **0.9639** |
| Avl | **1.000** | **0.969** | **0.9945** | **0.969** | **0.918** | **0.9600** |
| Avr | **0.978** | **0.956** | **0.9741** | **0.957** | **0.935** | **0.9531** |
| I | **1.000** | **0.963** | **0.9935** | **0.970** | **0.941** | **0.9649** |
| II | **0.985** | **0.961** | **0.9808** | **0.969** | **0.924** | **0.9611** |
| III | **0.982** | **0.958** | **0.9778** | **0.975** | **0.947** | **0.9701** |
| V1 | **0.985** | **0.959** | **0.9804** | **0.974** | **0.935** | **0.9671** |
| V2 | **0.998** | **0.960** | **0.9913** | **0.968** | **0.930** | **0.9613** |
| V3 | **0.995** | **0.971** | **0.9908** | **0.965** | **0.921** | **0.9572** |
| V4 | **0.985** | **0.966** | **0.9816** | **0.975** | **0.933** | **0.9676** |
| V5 | **1.000** | **0.971** | **0.9949** | **0.969** | **0.921** | **0.9605** |
| V6 | **0.989** | **0.967** | **0.9851** | **0.969** | **0.917** | **0.9598** |
| I, II, III, Combined | **1.000** | **0.973** | **0.9952** | **0.982** | **0.958** | **0.9778** |
| 12 leads combined | **1.000** | **0.998** | **0.9996** | **0.987** | **0.964** | **0.9829** |

**Table 2. Sensitivity of dynamic method for MI multiclass classification in the training dataset.**

| Leads | Sensitivity | | | | | | Accuracy |
|---|---|---|---|---|---|---|---|
|  | MI | VHD | DY | BBB | CH | HC |  |
| AVF | **0.998** | **0.825** | **0.383** | **0.368** | **0.539** | **0.981** | **0.896** |
| AVL | **0.997** | **0.767** | **0.653** | **0.473** | **0.624** | **0.984** | **0.906** |
| AVR | **1.000** | **0.836** | **0.267** | **0.528** | **0.567** | **0.973** | **0.913** |
| I | **0.996** | **0.782** | **0.775** | **0.336** | **0.629** | **0.986** | **0.939** |
| II | **0.998** | **0.656** | **0.326** | **0.316** | **0.482** | **0.985** | **0.899** |
| III | **0.995** | **0.848** | **0.742** | **0.819** | **0.696** | **0.975** | **0.931** |
| V1 | **0.996** | **0.848** | **0.865** | **0.827** | **0.912** | **0.984** | **0.939** |
| V2 | **0.997** | **0.784** | **0.787** | **0.834** | **0.927** | **0.997** | **0.942** |
| V3 | **0.998** | **0.858** | **0.878** | **0.823** | **0.918** | **1.000** | **0.949** |
| V4 | **1.000** | **0.824** | **0.797** | **0.817** | **0.906** | **0.986** | **0.936** |
| V5 | **0.998** | **0.818** | **0.839** | **0.862** | **0.897** | **0.998** | **0.927** |
| V6 | **0.996** | **0.782** | **0.822** | **0.816** | **0.912** | **0.987** | **0.928** |
| I, II, III Combined | **1.000** | **0.867** | **0.863** | **0.913** | **0.928** | **1.000** | **0.948** |
| 12 leads combined | **1.000** | **0.892** | **0.878** | **0.957** | **0.955** | **1.000** | **0.963** |

**Table 3. Sensitivity of dynamic method for MI multiclass classification in the test dataset.**

| Leads | Sensitivity | | | | | | Accuracy |
|---|---|---|---|---|---|---|---|
| | MI | VHD | DY | BBB | CH | HC | |
| AVF | 0.993 | 0.625 | 0.083 | 0.077 | 0.409 | 0.981 | 0.8246 |
| AVL | 0.994 | 0.667 | 0.500 | 0.333 | 0.500 | 0.960 | 0.8247 |
| AVR | 0.995 | 0.636 | 0.167 | 0.250 | 0.500 | 0.963 | 0.8134 |
| I | 0.995 | 0.632 | 0.500 | 0.231 | 0.529 | 0.981 | 0.8097 |
| II | 0.993 | 0.556 | 0.120 | 0.222 | 0.472 | 0.981 | 0.8096 |
| III | 0.994 | 0.818 | 0.542 | 0.789 | 0.680 | 0.975 | 0.8706 |
| V1 | 0.993 | 0.778 | 0.765 | 0.810 | 0.900 | 0.996 | 0.9067 |
| V2 | 0.994 | 0.824 | 0.737 | 0.800 | 0.890 | 0.995 | 0.9029 |
| V3 | 0.996 | 0.778 | 0.778 | 0.823 | 0.910 | 0.995 | 0.8992 |
| V4 | 0.993 | 0.824 | 0.737 | 0.797 | 0.920 | 0.988 | 0.8955 |
| V5 | 0.994 | 0.778 | 0.789 | 0.850 | 0.930 | 0.995 | 0.8993 |
| V6 | 0.993 | 0.782 | 0.812 | 0.842 | 0.909 | 0.979 | 0.9179 |
| I, II, III Combined | 0.996 | 0.727 | 0.286 | 0.811 | 0.425 | 0.987 | 0.9284 |
| 12 leads combined | 0.998 | 0.812 | 0.748 | 0.926 | 0.955 | 0.997 | 0.9483 |

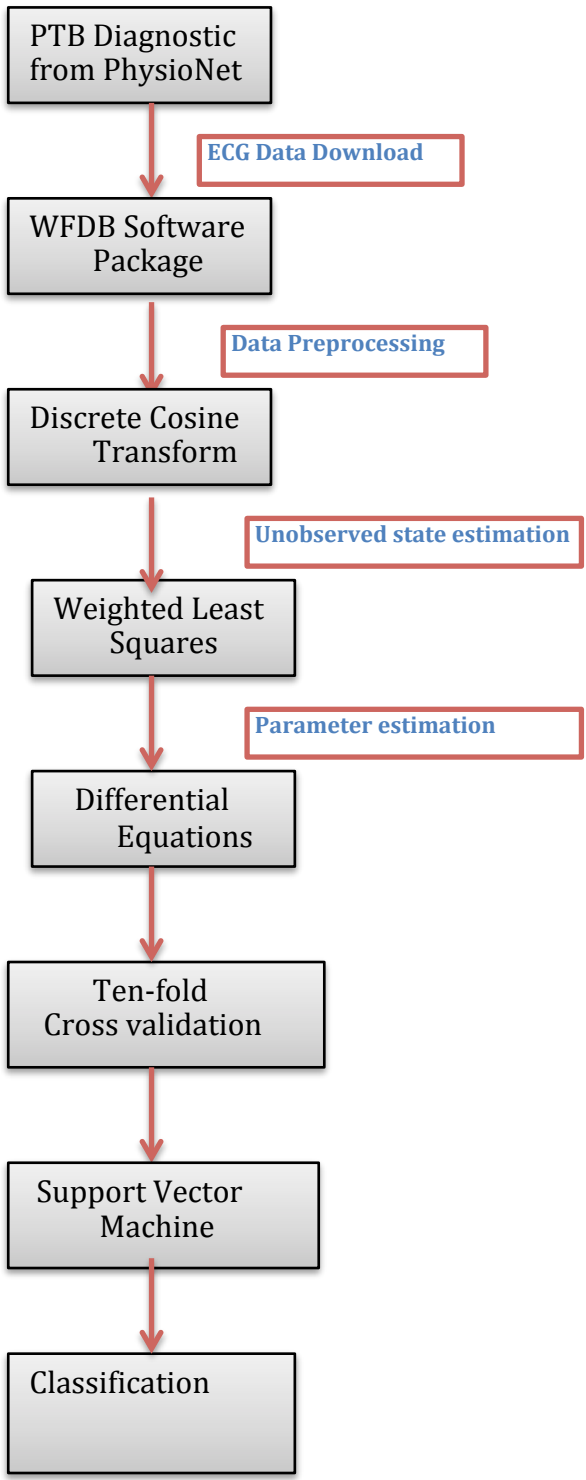

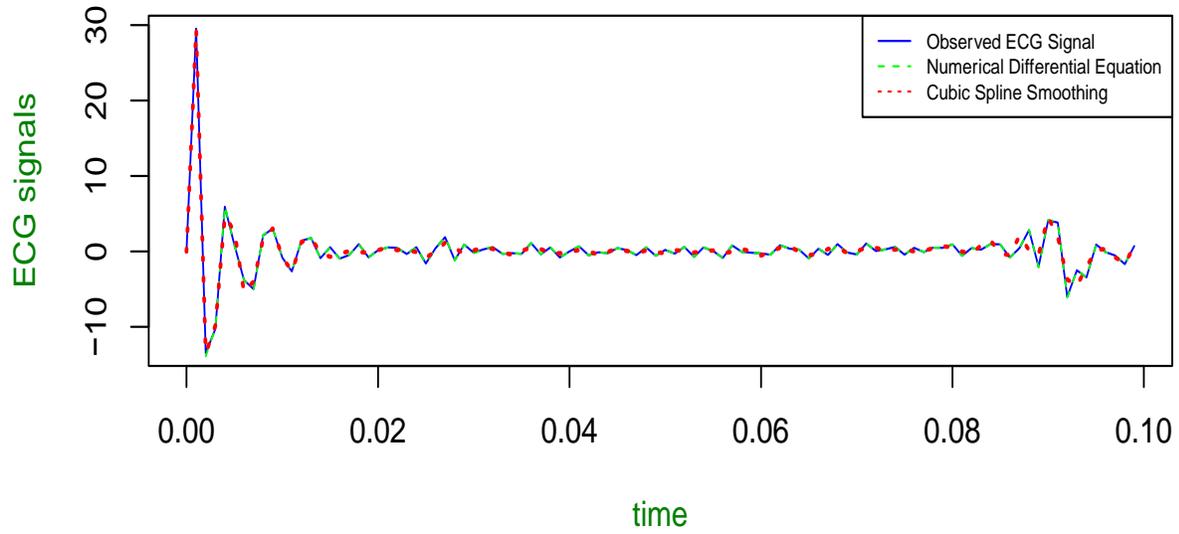

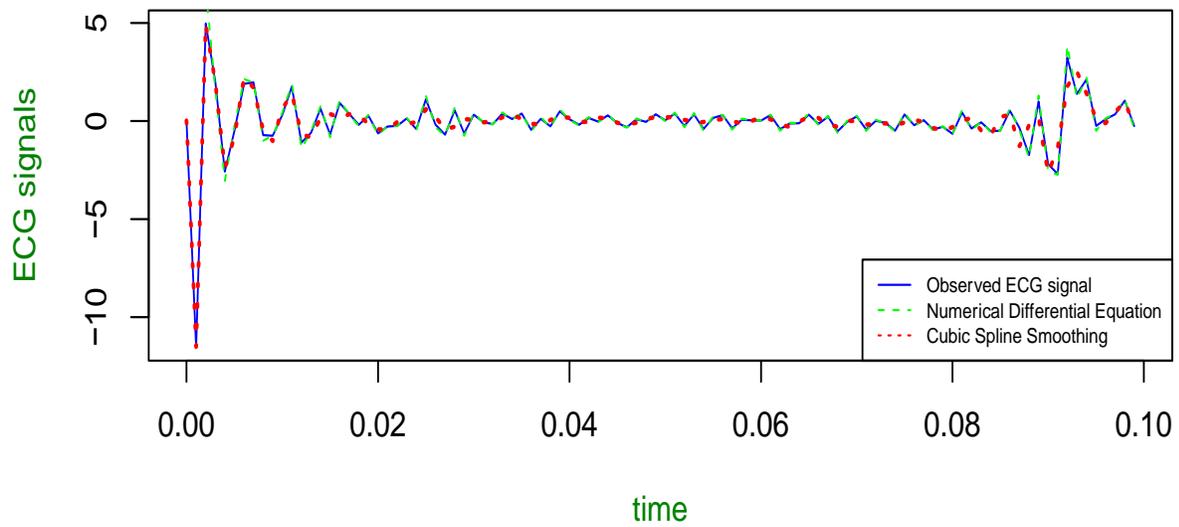

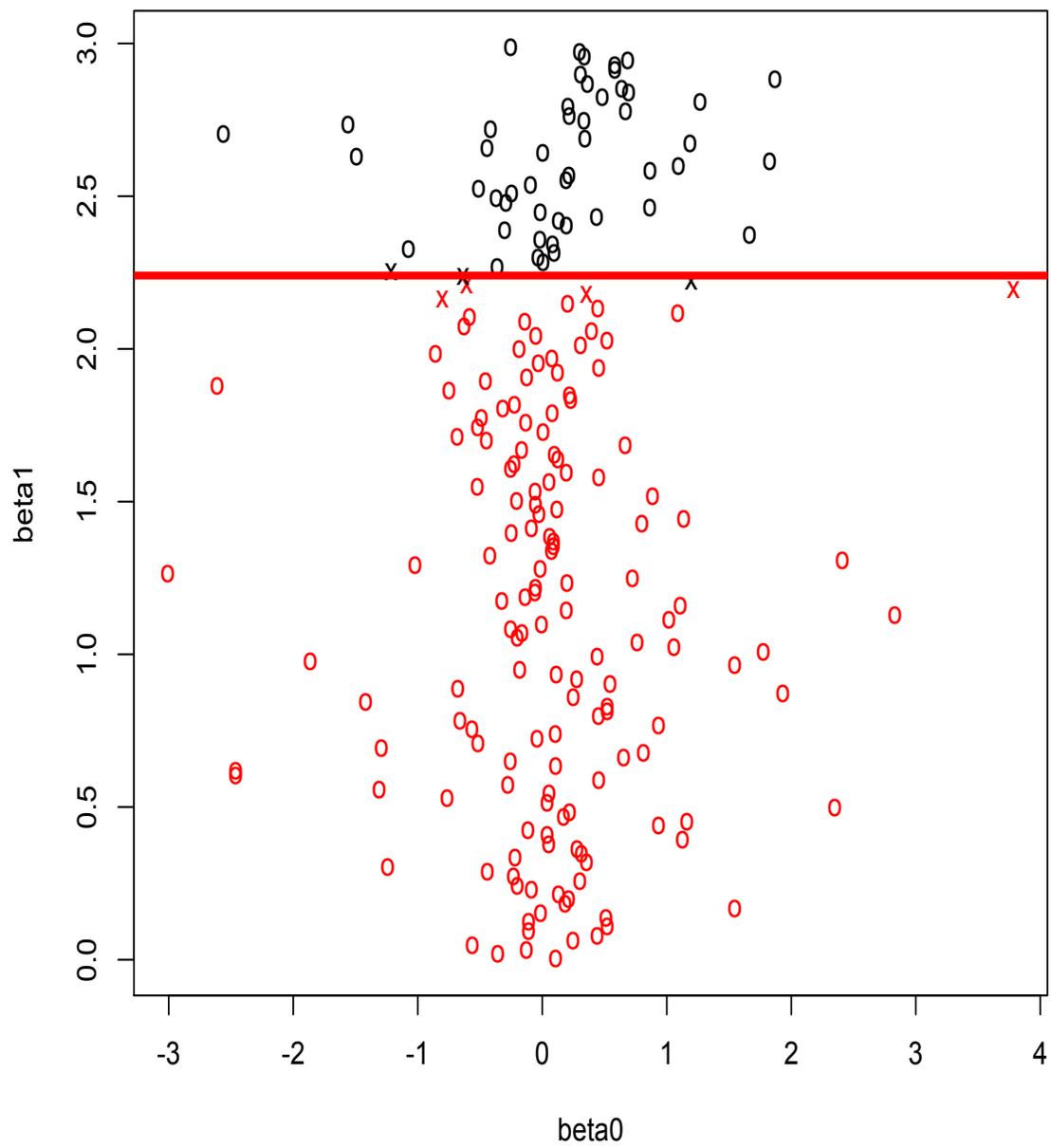

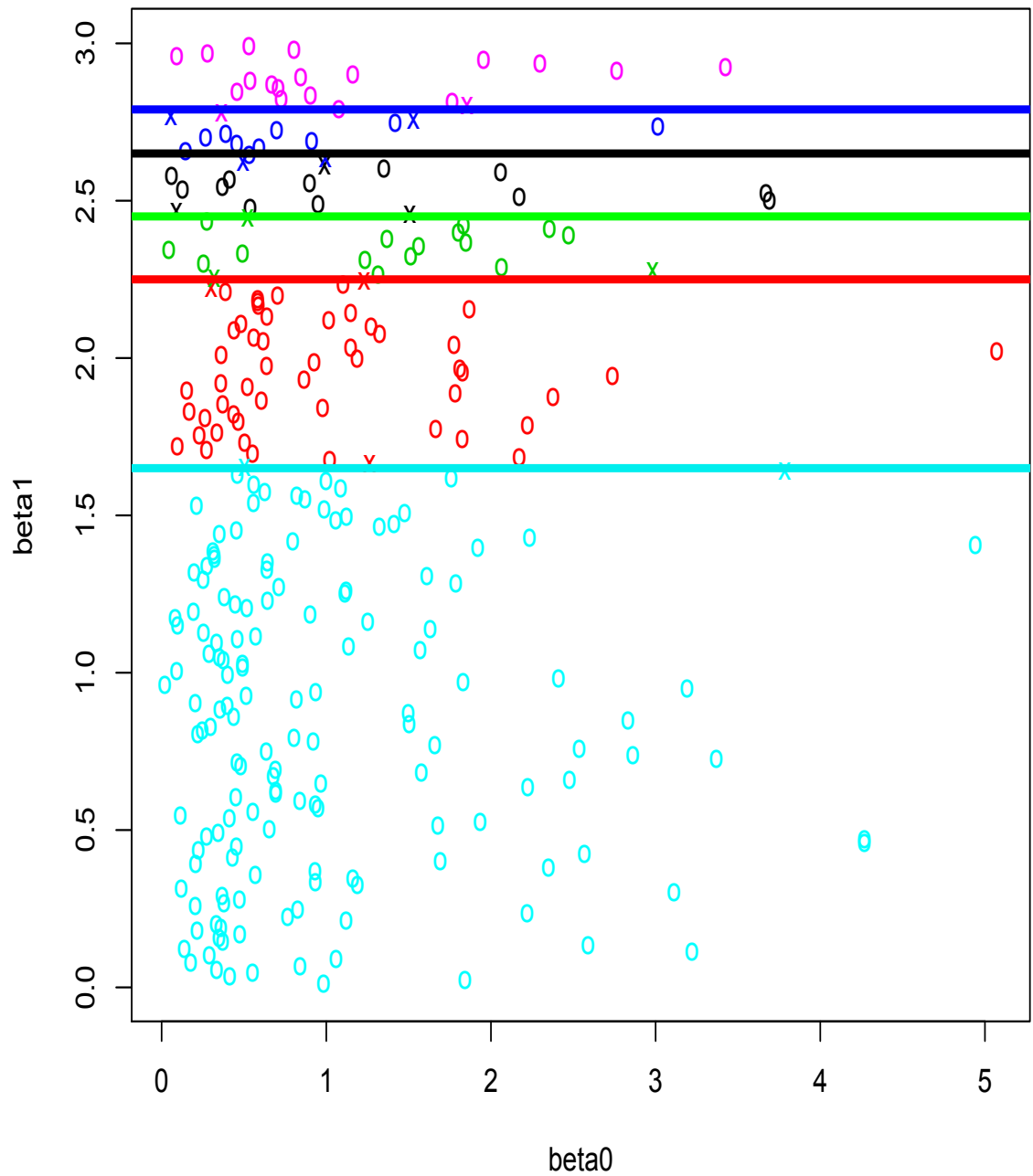